\pdfoutput=1
\documentclass[letter,oneside]{article}
\usepackage[margin=1.0in]{geometry}

\usepackage[numbers]{natbib}

\usepackage{amsmath,amsfonts,bm}

\def\eqref#1{equation~\ref{#1}}

\def\1{\bm{1}}

\DeclareMathAlphabet{\mathsfit}{\encodingdefault}{\sfdefault}{m}{sl}
\SetMathAlphabet{\mathsfit}{bold}{\encodingdefault}{\sfdefault}{bx}{n}

\DeclareMathOperator*{\argmax}{arg\,max}

\usepackage{hyperref}
\usepackage{url}

\usepackage{times}
\usepackage{latexsym}
\usepackage{graphicx}

\usepackage{amsmath,amssymb,amsfonts,amsthm}
\usepackage{algorithm, algorithmic}

\usepackage{multicol,multirow,booktabs}

\usepackage{wrapfig}
\usepackage{color}
\definecolor{darkgreen}{rgb}{0.1, 0.5, 0.1}
\newcommand{\eat}[1]{}

\usepackage{mdframed}
\usepackage{listings}
\definecolor{postit}{RGB}{254,255,195}
\title{\textbf{\texttt{CELL}} your Model: Contrastive Explanations for Large Language Models}

\author{Ronny Luss \quad \quad Erik Miehling \quad \quad Amit Dhurandhar \\
IBM Research \\
\texttt{rluss@us.ibm.com, erik.miehling@ibm.com, adhuran@us.ibm.com}}
\date{}

\begin{document}

\maketitle

\begin{abstract}
The advent of black-box deep neural network classification models has sparked the need to explain their decisions. However, in the case of generative AI, such as large language models (LLMs), there is no class prediction to explain. Rather, one can ask why an LLM output a particular response to a given prompt. In this paper, we answer this question by proposing a contrastive explanation method requiring simply black-box/query access. Our explanations suggest that an LLM outputs a reply to a given prompt because if the prompt was slightly modified, the LLM would have given a different response that is either less preferable or contradicts the original response. The key insight is that contrastive explanations simply require a scoring function that has meaning to the user and not necessarily a specific real valued quantity (viz. class label). To this end, we offer a novel budgeted algorithm, our main algorithmic contribution, which intelligently creates contrasts based on such a scoring function while adhering to a query budget, necessary for longer contexts. We show the efficacy of our method on important natural language tasks such as open-text generation and chatbot conversations.\end{abstract}

\section{Introduction}
Generative artificial intelligence (AI) has rapidly transformed society and will continue to do so for the foreseeable future, albeit in ways we do not yet know. Thusfar, it has impacted how people conduct their jobs (e.g., code generation for software engineers \citep{codegenerationai}, text summarization for lawyers \citep{lexislegalai} and doctors \citep{healthcareai}) as well as how people conduct their daily activities (e.g., rewriting emails, seeking advice, or designing vacation itineraries). As AI has advanced over the last two decades, so did the need for explaining how the AI was making decisions (e.g., why was a customer denied a bank loan or why was an image classified as a pedestrian crossing a street). Such explanations have been the topic of regulations in the USA with the AI Bill of Rights \citep{AIBillofRights} and in Europe with the GDPR \citep{gdpr} and the recent EU AI Act \citep{euaiact}.\eat{Such needs will further prove critical for advancements resulting from the recent massive investment announcement in AI infrastructure in the USA \citep{stargate}.} 

Much has been done in explainable AI typically regarding classification and regression (see surveys \citet{guidotti2018survey} and \citet{yang2023survey}) mostly focusing on black-box models, e.g., deep neural networks. Explanation methods vary and include attribution methods such as LIME \citep{lime}, SHAP \citep{unifiedPI}, and saliency \citep{saliency}, and contrastive explanations such as CEM \citep{CEM} and CAT \citep{cat}.  

\begin{figure*}[t]
    \centering
    \includegraphics[width=\textwidth]{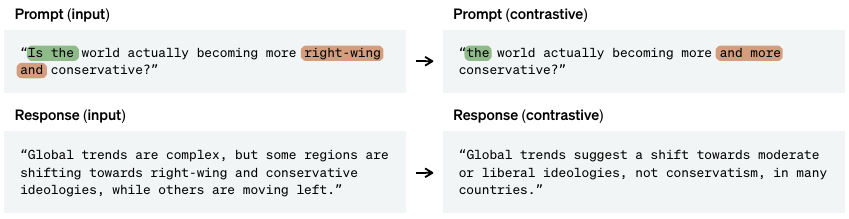}%
    \vspace{0.3em}
    \hrule\hrule
    \vspace{0.2em}
    \includegraphics[width=\textwidth]{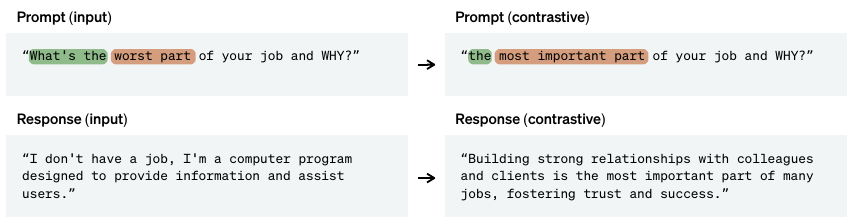}
    \vspace{-0.5cm}
    \caption{Contrastive explanations for natural language generation by \texttt{meta-llama/Llama-3.1-8B-Instruct}. Colors match what is changed between input and contrastive prompts. These explanations suggest that the input prompt generated the input response because if the highlighted changes were made, the new contrastive prompt would generate a different response which contradicts the input response. Prompts taken from the Moral Integrity Corpus \citep{micdata}.}
    \label{fig:mic_intro}

\end{figure*}

The focus of this paper is contrastive explanations for Large Language Models (LLMs). In the typical classification setting, contrastive explanations dictate that a classifier predicted label $y$ on sample $x$ because if $x$ was slightly modified to be $x_c$, the classifier would have predicted label $y_c$ instead. In the case of LLMs, there is no classifier and the output is a sequence of words. While explainability is a well-studied area for classifiers, explanations for LLMs are still limited. A recent method,  MExGen \citep{mexgen}, derives LIME and SHAP methods for LLMs based on mask infilling while TextGenSHAP \citep{textgenshap} speeds up SHAP for LLMs using ``speculative decoding", but these are attribution methods which highlight words in the prompt to maintain the response (not change it) also requiring real valued representations of the response. Such attribution methods are complimentary to our proposal of a contrastive method; they are more restrictive as they can explain only individual tokens or need more information when explaining entire responses. 

\eat{To the author's knowledge, this paper offers the first contrastive explanation methods for LLMs. }Consider the examples in Figure \ref{fig:mic_intro}. Given an input prompt that is fed to an LLM, we ask why the LLM output a particular response. Our methods create perturbations of the input prompt, called contrastive prompts, which when fed to the LLM result in contrastive responses that differ from the input response in some user defined manner, e.g., a contrastive response that contradicts the input response. In the top example, the contrastive explanation dictates:  the LLM responded by suggesting that political trends are complex because if the prompt had asked solely about conservatism rather than \emph{right-wing} conservatism, it would have responded with the opposite answer. The bottom example showcases the effects of guardrails; asking about the \emph{most important} rather than the \emph{worst} part of the LLM's job allowed it to give an informative answer. Additional examples are in the Appendix Section \ref{s:additional_qualitiative_examples}.

\emph{The key insight here is that contrastive explanations simply require a scoring function that a user can interpret and not necessarily a specific class label.} Moreover, given that input prompts may have large contexts (viz. in Retrieval Augmented Generation (RAG)), we propose an approach that can efficiently find contrasts with a limited number of calls to the black-box model, something that is not considered in previous works. 

\noindent{\textbf{Remarks on previous contrastive explanations.}} Various works on contrastive explanations are discussed below in Section \ref{s:related-work}; they are all designed for classification problems such as sentiment analysis or natural language inference, whereas our work applies to language generation. The difference between explaining the response of an LLM to an input text and that of a classifier to the same input text is subtely different. While, in classification, the explanation is a function of a modified prompt (or text), in the case of language generation, the explanation has to consider the triplet of the modified prompt, its resulting response, and the original response, as depicted in Figure \ref{fig:cem-llm-budget} and further discussed in Section \ref{ss:scoring_functions}. In essence, the contrastive method has to work with a scoring function that considers both the original and the modified response, and not just a real valued label as in classification. Nevertheless, our first algorithm, \textbf{\texttt{m-CELL}} in Section \ref{ss:cell}, can be viewed as an extension of recent related works on classification such as \cite{cat}, while our second algorithm, \textbf{\texttt{CELL}} in Section \ref{ss:cell-budget}, provides a novel adaptive search for contrastive explanations highly relevant to the generative setting where prompts and responses can be large. Another related group of works seek to prompt LLMs to generate contrastive explanations; these works have also only been designed for classification, but we adapt a version as a baseline in our experiments.

\noindent{\textbf{Contributions.}} We propose two methods to generate contrastive explanations for LLMs: a myopic algorithm that is effective for small prompts and a budgeted algorithm that scales for large contexts. We demonstrate quantitatively that these algorithms are effective and practical. Finally, we apply contrastive explanations to a real application involving conversations, showcasing their practicality. 

\section{Related Work}
\label{s:related-work}
\eat{\citet{xainlp_survey} considered explainability for natural language processing, primarily for classification tasks where local explanations were provided, among which our focus is on post-hoc methods that explain a fixed model's prediction. One large group of explainability methods are feature based where the explanation outputs some form of feature importance (i.e., ranking, relevance, etc.) of the words in text \citep{dknn_text,dknn,inputreduction,patternattribution_text,lime,melis_jaakkola_2017}. Other types of local post-hoc explanations include exemplar based \citep{proto,infl,l2c} that output similar instances to the input.}

Among local methods, our focus is on contrastive methods \citep{cat, CEM, gyc, CEM-MAF, mice}. Contrastive explanations are complementary to attribution and exemplar style explanations \citep{arya2019explanation} (discussed in Appendix Section \ref{s:additional_related_work}) as they provide ways to realistically manipulate the input in a minimal manner so as to change the output. In our setup, we want to modify the input prompt so that an LLM produces an output with a different user specified quality or characteristic (viz. fairness, preference, etc.). The latter distinguishes our work from prior contrastive explanations for the classification setting. 

Another contrastive method POLYJUICE \citep{polyjuice} is a human-in-the-loop method requiring supervision about the type of edit to be performed to the text such as negation, word replacement, etc. A contrastive latent space method \citep{jacovi} does not generate contrastive text, but rather highlights (multiple) words in the input text that are most likely to alter a classification prediction, and is furthermore not a black-box method. Similarly, \citet{yin2022} highlight words that influence a model predicting a target output instead of a \emph{foil}; this work is related to saliency and uses gradient-based scoring. A few recent works use LLMs to generate contrastive explanations \citep{core,disco,promptingLLM2024,madsen-etal-2024-self} but focus on classification. 
\section{Formulation}
We here formulate the contrastive explanation problem for LLMs. Denote by $x_0$ an input prompt and $\mathcal{X}$ the space of prompts, i.e., strings. Let $LLM(x)$ be the response of an LLM to prompt $x$. Define $g(x_0, y_0, y_c)$ as a scoring function that inputs a prompt $x_0$, the initial response $y_0=LLM(x_0)$, a perturbed version of $x_0$ denoted as $x_c$, and the response to another prompt $x_c$ denoted as $y_c=LLM(x_c)$ where $x_c$ is a typically a perturbed version of $x_0$. Also denote $f(x_0, x_c)$ as a measure of similarity between two prompts $x_0$ and $x_c$. We formulate the contrastive explanation problem for LLMs as
\begin{align}
\label{eq:cem-llm}
\mbox{minimize} & \quad f(x_0,  x)\\
\mbox{subject to} &\quad  g\left(x_0, LLM(x_0), LLM(x)\right) \ge \delta \nonumber \\
 & \quad x \in \mathcal{X} \nonumber
\end{align}
Assuming bounded $\mathcal{X}$, Problem (\ref{eq:cem-llm}) is a combinatorial optimization problem over all possible prompts in $\mathcal{X}$. Note that this generalizes contrastive explanations \citep{CEM} or adversarial attacks \citep{carlini, l1adv} where typically $LLM(\cdot)$ is replaced by a classifier, $g(\ldots)$ does not depend on $x_0$, and the constraint is such that the predicted class of $x_0$ changes. Contrastive explanation methods further constrain the contrastive explanation, i.e., the solution to (\ref{eq:cem-llm}), to lie on a manifold that maintains it to be a realistic example. In the case of language generation, such constraints will be enforced by infilling masks, i.e., replacing missing word(s).

\noindent \textbf{Similarity:} Experiments in this paper measure prompt similarity $f(x_0, x_c)$ as the number of mask and infill operations applied to a string $x_0$ to obtain string $x_c$. Other functions could be considered based on commonly used text similarity metrics such as BLEU or ROUGE. Our choice is selected to focus on minimizing the number of LLM queries made by the algorithms described below. 

\eat{
In order to formalize how we solve problem (\ref{eq:cem-llm}) in practice, denote by $\mathcal{X}_n$ the space of ordered lists of $n$ strings. For example, the list \texttt{[``The boy ran", ``a race and", ``crossed the finish", ``line"]} $\in\mathcal{X}_4$. We define a function $\texttt{split\_prompt(x)}$  that take string $x$ and returns a list of substrings such that the concatenation of them is $x$. For example, given the string $x$\,=\,\texttt{``The boy ran a race and crossed the finish line"}, $\mbox{split\_prompt}(x)$ could result in the list above. Note the function \texttt{split\_prompt$(\cdot)$} can be chosen to output lists of unigrams or any other breakup of a string. Define $\mbox{mask}(x,Z,j)$ as a function that inputs a prompt $x$, list of substrings $Z$, and index $j$, and outputs a prompt with the $j^{th}$ substring in $Z$ that appears in $x$ replaced by a \texttt{<mask>} token. 
}

\subsection{Scoring Functions}
\label{ss:scoring_functions}
Problem (\ref{eq:cem-llm}) requires users to provide a scoring function for their particular usecase. We formalize scoring functions used in Section \ref{s:experiments}. \eat{Scoring functions below can depend on any subset of the inputs $x_0, y_0,\mbox{and } y_c$, as defined above. }It is important to note that these user-defined scoring functions need not be symmetric and can incorporate direction. For example, the preference score defined below incorporates direction (whether preference increases or decreases) but can also be defined by the absolute value of the score instead.

\noindent \textbf{Preference:} This scoring function outputs a score defining which of two responses is \textit{preferable} for a given prompt. Specifically, we use the \path{stanfordnlp/SteamSHP-flan-t5-xl} LLM available on HuggingFace \citep{steamshp} which is trained to predict how helpful each response is for the prompt. These scores are normalized to act as probabilities, and our preference score is the difference between the two probabilities. \eat{Such a scoring function can be used for explaining natural language generation.}

\noindent \textbf{Contradiction:} This scoring function inputs two responses $y_1$ and $y_2$ obtained from an LLM. A Natural Language Inference (NLI) model is used to score the likelihood that $y_1$ contradicts itself (as a reference likelihood), denoted as $p_1$, and the likelihood that $y_2$ contradicts $y_1$, denoted as $p_2$. We define the contradiction score as the difference $p_2-p_1$. Note that $p_1$ should be small for a good NLI model but is still computed here to give a reference point. Experiments in this paper use the NLI model \path{cross-encoder/nli-deberta-v3-base} available on HuggingFace. \eat{Such a scoring function can also be used for explaining natural language generation.}

\noindent \textbf{BLEU\_SUMM:} The BLEU score, between 0 and 1, measures the similarity of two text samples (closer to 1 is more similar). Given two prompts $x_1$ and $x_2$ and their corresponding responses $y_1$ and $y_2$, we measure the BLEU score between prompts $x_1$ and $x_2$, denoted as $a$, and the BLEU score between responses $y_1$ and $y_2$, denoted as $b$, and assign a score as $w_1\cdot a + w_2\cdot(1-b)$, meaning a higher score is given for having similar prompts and dissimilar responses. Our experiments use $w_1=0.2, w_2=0.8$ to give more importance to dissimilarity of responses.

\begin{figure*}[t]
    \centering
    \includegraphics[width=\textwidth]{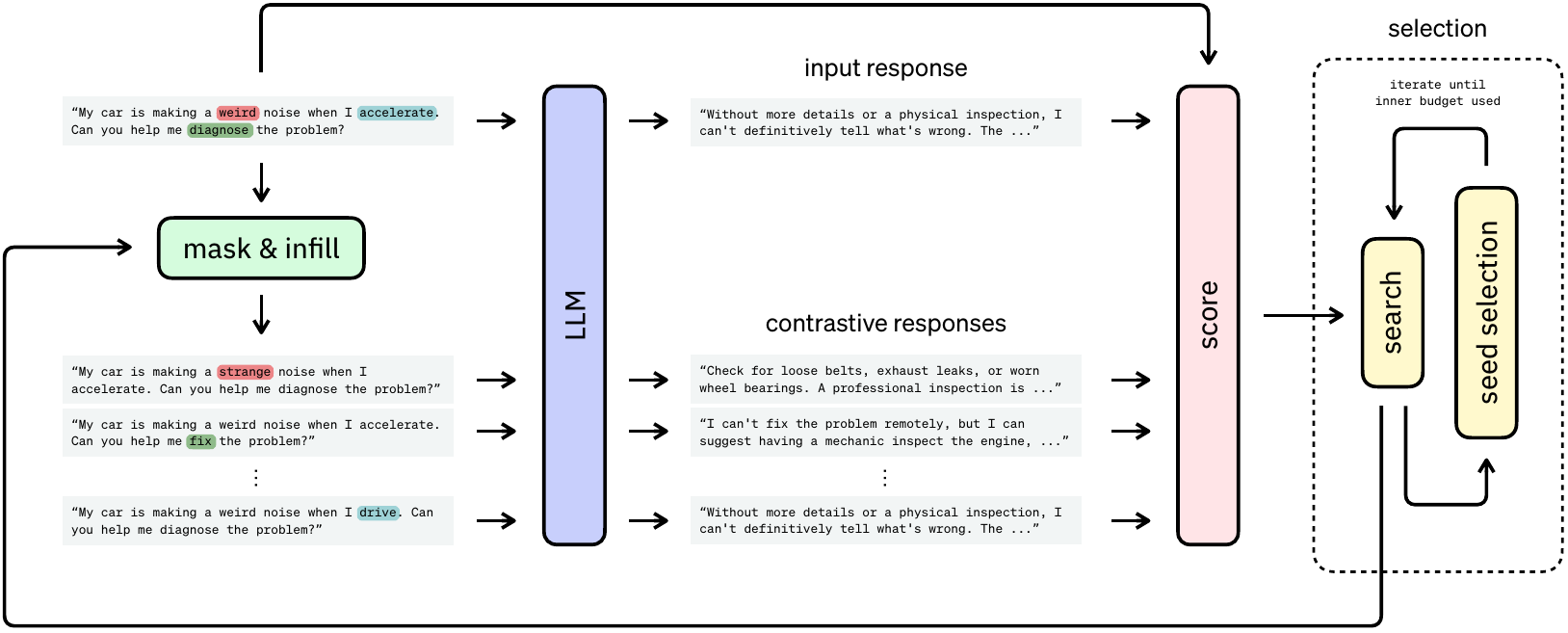}
    \vspace{-0.5cm}
    \caption{Illustration of the \textbf{\texttt{CELL}} and \textbf{\texttt{m-CELL}} algorithms. Both algorithms can be summarized as an iterative process that repeats a) Select substrings of the prompt to search, b) Generate perturbed prompts (mask \& infill), c) Generate responses for each perturbed prompt (via the LLM), d) Score each perturbed prompt/response. The main difference between the budgeted method, \textbf{\texttt{CELL}}, and the myopic method, \textbf{\texttt{m-CELL}}, is in the \emph{selection} block -- \textbf{\texttt{CELL}} augments the search process with a prompt seed generation step (see Algorithm \ref{algo:cem_llm_budget_summary} for details). \textbf{\texttt{CELL}}'s search is an iterative loop subject to an inner loop budget before repeating the prompt seed generation step, whereas the myopic method's search simply enumerates over substrings.}
    \label{fig:cem-llm-budget}
\end{figure*}

\begin{algorithm}[t!]
\caption{{\bf \texttt{CELL}}}
  \label{algo:cem_llm_budget_summary} 
\begin{algorithmic}
\STATE \textbf{Input:} $LLM(\cdot)$, scoring function $g(\cdots)$, infiller $I(\cdot)$, threshold $\delta$, prompt $x_0$, budget $B$, max iters $T$, prompt seed ratio $\alpha$, and let $q = \lfloor B / \log(B)\rfloor$

\STATE  $Z \leftarrow  \mbox{split\_prompt}(x_0)$  \# Get substrings to be masked

\STATE  $X \leftarrow \{\}$ $\quad$ \# Keep track of perturbed prompts

\FOR{$t=1$ to $T$}
    \STATE $n_c \leftarrow \textbf{\texttt{NUM\_SEEDS}}(t, B)$ \# number prompt seeds
    \STATE  \# Generate $n_c$ perturbations as seeds to search from: 
    \STATE Set $n_1=\min(\alpha\cdot n_c, |X|)$ and $n_2=n_c-n_1$
    \STATE  \textbf{Prompt Seeds $x_0$}: Select $n_2$ substrings from $Z$ and generate perturbations $X_1$ by infilling with $I(\cdot)$.
    \STATE \textbf{Prompt Seeds $x_0$ perturbed}: Select $n_1$ prompts from $X$ and generate perturbations $X_2$ using $I(\cdot)$.  
    \STATE $X_C \leftarrow X_1 \cup X_2$ $\quad$ \# Current seed prompt samples
    \STATE $X \leftarrow X \cup X_C$ \# All generated perturbations
    \STATE $m \leftarrow n_c$
    \FOR{$j=1$ to $\lceil \log(n_c) \rceil$}
        \STATE $n_p \leftarrow \left\lfloor q/(m\lceil\log(n_c)\rceil) \right\rfloor$ \# Num. prompts per seed
        \STATE  $X_p \leftarrow \textbf{\texttt{SAMPLE\_SEEDS}}(X_C, Z, n_p, I(\cdot))$
        \STATE \# Score all perturbed prompts in $X_p$
        \FOR{$x\in X_p$}
            \STATE  Compute $LLM(x)$ and scores using $g(\cdots)$
            \STATE $n_b \leftarrow n_b + 1$ $\quad$ \# Number LLM inferences made
            \IF{$n_b \ge B$}
                \STATE \textbf{Output: }Best perturbed prompt/response
            \ENDIF
        \ENDFOR
        \STATE $X \leftarrow X \cup X_p$
        \IF{\normalfont{Best score found is greater than} $\delta$}
           \STATE  \textbf{Output: }Best perturbed prompt/response
         \ENDIF
        \STATE $m \leftarrow \lceil m/2 \rceil$
        \STATE $X_C \leftarrow \textbf{\texttt{BEST\_SUBSET}}(X_p, m)$
    \ENDFOR
\ENDFOR
\end{algorithmic}
\end{algorithm}

\section{Methods}
In this section, we describe two variants of our
\underline{c}ontrastive \underline{e}xplanation method for \underline{l}arge \underline{l}anguage models ({\bf \texttt{CELL}}) for searching the space of contrastive examples. In practice, this is done by splitting a prompt into $n$ substrings and searching over the space of all possible masked and infilled subsets of these $n$ substrings. 
The first algorithm, \textbf{\texttt{m-CELL}}, is a myopic search over potential substrings to replace, and can be viewed as an extension of previous contrastive explanations for classifiers. The second algorithm, \textbf{\texttt{CELL}}, is our main algorithmic contribution and involves an adaptive 
search 
constrained by a budget on the number of calls to the LLM being explained. Such calls can become expensive due to long documents (e.g., as with text summarization tasks). A key novelty over previous contrastive explanations (for classifiers), such as \citet{cat}, \citet{CEM}, and \citet{gyc}, is the insight to use scoring functions that relate the input prompt to responses generated by modified prompts; this is the essence of defining contrastive explanations for a generator, such as an LLM, versus a classifier. 

Both methods require the following inputs: an LLM to be explained, a scoring function $g(\cdot, \cdot, \cdot)$ as defined above, and an infiller $I(\cdot)$ that receives as input a string with a \texttt{<mask>} token and outputs a string with \texttt{<mask>} replaced by new text. Various options exist for the infiller model; these include BERT-based models that replace \texttt{<mask>} with a single word, and BART or T5-based models that replace \texttt{<mask>} with potentially multiple words (allowing for addition or deletion of words in addition to simple substitution). While finetuning an infiller could further help performance (e.g., as \citet{mice} do for classification), this is expensive and out of the scope of the current work. 

Figure \ref{fig:cem-llm-budget} illustrates the general logic common to both methods. 
Specifically, at each iteration, a list of perturbed prompts are selected and passed to the infiller to generate new perturbed prompts. These prompts are then passed through the LLM to generate corresponding responses. A task-dependent score is computed based on the input prompts and response and the perturbed prompts and responses (or any subset of these prompts and responses). Lastly, the score is used to determine which perturbed prompts to continue searching from until a sufficiently modified contrastive prompt is found.

\textbf{\texttt{CELL}} and \textbf{\texttt{m-CELL}} split prompts into substrings of consecutive words. Setting \texttt{split\_k=1} splits prompts into individual words. Setting \texttt{split\_k=2} splits prompts into consecutive pairs of words, and so forth. Hence higher \texttt{split\_k} results in a smaller search space.

\subsection{\texttt{m-CELL}}
\label{ss:cell}
Our myopic search, \textbf{\texttt{m-CELL}}, uses the following strategy: An input prompt is first split into $n$ substrings (according to \texttt{split\_k}); the contrastive example will be a perturbed prompt that masks and replaces a subset of these $n$ substrings. In the first iteration, each of the $n$ substrings is masked and infilled, the $n$ perturbed prompts are passed through the LLM to generate $n$ responses, and these responses are used to compute $n$ scores. If a response results in a sufficiently large score, the corresponding perturbed prompt and response is deemed the contrastive example; otherwise, the perturbed prompt resulting in the largest score is used as the initial prompt and the same steps are followed on the $n-1$ remaining original substrings. These steps are repeated until either a contrastive example is found or all substrings have been perturbed without finding a contrastive example.  This strategy extends contrastive explanations of \citet{cat} for classifiers to language generation. Pseudocode for \textbf{\texttt{m-CELL}} can be found in Algorithm \ref{algo:cem_llm_greedy} in the Appendix.

\subsection{\texttt{CELL}}
\label{ss:cell-budget}
When the search is over a prohibitively large number of substrings, as is typical in text summarization for example, one might be conscientious of how many times the LLM is called. The next algorithm, called \textbf{\texttt{CELL}}, explores new perturbations from the input prompt while also exploiting perturbations already made. This algorithm, detailed in Algorithm \ref{algo:cem_llm_budget_summary}, is inspired by \citet{trustregionexpl2024} which adaptively samples a continuous search space subject to a budget; their task is to find a trust region that satisfies local explainability properties whereas our task is to find a region that satisfies a score criterion. 

Each iteration is broken down into three main blocks: 1) Compute the number of seeds, i.e., prompts, to perturb, 2) Generate seeds, and 3) Search around these seeds (inner loop). Each iteration of the inner loop samples a particular number of prompts around those seeds in order to use the total budget. Function \textbf{\texttt{NUM\_SEEDS}} could take various forms; one such form (Algorithm \ref{algo:num_centers} in the Appendix) is inspired by optimal sampling from continuous distributions.

Our method deviates from \citet{trustregionexpl2024} as it searches over a discrete space. \textbf{\texttt{CELL}} employs a seed-driven approach 
where some seeds are generated from the initial prompt and others 
from previously perturbed prompts. This allows the search to explore new perturbations of the initial prompt while also taking advantage of favorable perturbations previously made (hyperparameter $\alpha$ controls the tradeoff). The search around prompt seeds in the inner loop first samples a fixed number of perturbations around each seed using function \textbf{\texttt{SAMPLE\_SEEDS}} (Algorithm \ref{algo:sample_centers} in the Appendix) and checks if a contrastive example was found. The next iteration of this inner loop reduces the number of seeds sampled from the current list of perturbed prompts and increases the number of samples taken around each seed. The decrease/increase in seeds/samples focuses more on perturbations more likely to lead to contrastive examples. Function \textbf{\texttt{BEST\_SUBSET}} outputs prompt seeds as ordered by $g(\cdot, \cdot, \cdot)$.

\section{Experiments}
\label{s:experiments}
LLMs have previously been used to explain classifiers by prompting for contrasts \cite{disco}.  Explaining LLMs through contrastive explanations is a different task and a novel direction for LLM explainability. We show how \textbf{\texttt{CELL}} performs against a baseline, described below, using an LLM-as-a-Judge framework. Additionally, we demonstrate \textbf{\texttt{CELL}} across several performance measures that demonstrate explanation properties: preference, edit distance, flip rates, and number of model calls. We conclude with an application to explaining conversations. Experiments were conducted with 1 \texttt{A100\_80gb} GPU and up to 64 GB memory.

\noindent{\textbf{Datasets and Models:} } We consider two datasets for the following experiments: the Moral Integrity Corpus (MIC) \citep{micdata} (using 500 prompts) and the Extreme Summarization (XSum) dataset \citep{xsum-emnlp} (using 500 documents). Two LLMs are used: \path{meta-llama/Llama-3.1-8B-Instruct} and \path{meta-llama/Llama-3.1-70B-Instruct}. Infilling is done using \texttt{T5-large} (finetuning an infiller is expensive and out of the scope of the current work). All corresponding standard error tables are in the Appendix.

\noindent{\textbf{Baseline:}} We take inspiration from classifier explanations that prompt LLMs to modify text such that the modified text maintains a particular classification \cite{disco}. We extend such a concept, referred to as \textbf{\texttt{Baseline}} in all following experiments, with a two-step procedure: 1) prompt the LLM (\path{meta-llama/Llama-3.1-8B-Instruct} in the following experiments) to \texttt{"Modify the following prompt with as few edits as possible such that it should elicit a response that contradicts the following response: RESPONSE"}, where \texttt{RESPONSE} is the LLM's response to the input prompt, 2) pass the modified prompt through the LLM being explained to generate a contrastive response. For experiments with a scoring function, the prompt is adapted with \texttt{"\ldots elicit a response that is less preferable than the following response: RESPONSE"}.

\subsection{LLM-as-a-Judge Comparisons}
\label{ss:llm_as_a_judge}
In deference to expensive human evaluations, we go the direction of using an LLM to evaluate explanations. Previous studies on contrastive explanations for classifiers \cite{cat} used a task to evaluate whether humans could predict class based on the information from contrastive explanations. For \textbf{\texttt{CELL}}, this would be akin to asking a human to predict the response of an LLM, which is the point of the explanation - humans cannot predict an LLM response. Furthermore, the contrastive explanation for LLMs outputs two pieces of information, a contrastive prompt and the corresponding response, and we thus want to make comparisons for both.

In this spirit, we ask an LLM to judge the contrastive prompt generated by \textbf{\texttt{CELL}} against \textbf{\texttt{Baseline}}, and further compare the corresponding responses using a classifier. Prometheus2 \cite{prometheus2} is a recent state-of-the-art LLM-as-a-judge trained to evaluate a given criterion among two responses. The judge is fed the criterion, \texttt{"Is the edited prompt elicited close, in terms of semantics, to the original prompt?"}, and outputs which contrastive prompt, that of \textbf{\texttt{CELL}} or of \textbf{\texttt{Baseline}}, is more closely aligned with the criterion. More details regarding the judge template are available in the Appendix. 

Table \ref{tab:llm_as_a_judge} shows results for explaining \path{meta-llama/Llama-3.1-8B-Instruct} responses to prompts from the MIC dataset. LLM temperature is varied to show robustness of the results. The percentage of \textbf{\texttt{CELL}} prompts judged to be more favored is significantly more than for \textbf{\texttt{Baseline}}. The numbers do not sum to 100\% because the judge sometimes does not favor one over the other. Furthermore \textbf{\texttt{CELL}} finds contrasts at a high rate, while \textbf{\texttt{Baseline}} performs poorly in this respect. Together, these observations show that \textbf{\texttt{CELL}} robustly outperforms \textbf{\texttt{Baseline}}.

\begin{table}[t]
\small
\centering
\caption{Results using Prometheus2 to judge \textbf{\texttt{CELL}} vs \textbf{\texttt{Baseline}} prompts and an NLI classifer to judge responses. \textbf{\texttt{CELL}} generates more favorable prompts with far more contrasts.}
\label{tab:llm_as_a_judge}
\footnotesize
\setlength{\tabcolsep}{2.5pt}
\begin{tabular}{|c|c|c|c|c|c|c|c|}
\multicolumn{8}{c}{\textbf{Scoring function: Contradiction}} \\ \hline
\multicolumn{1}{|c}{} & \multicolumn{7}{c|}{\textbf{Judge Response}} \\ \hline
Temperature& 0.25 & 0.5 & 0.75 & 1.0 & 1.25 & 1.50 & 1.75 \\ \hline
\% \textbf{\texttt{CELL}} favored& \textbf{58.6} & \textbf{60.6} & \textbf{58.0}& \textbf{53.0} & \textbf{50.6} & \textbf{53.2} & \textbf{56.0} \\ \hline
\% Baseline favored & 36.6 & 36.4 & 39.2 & 44.6 & 47.6 & 44.8 & 41.6 \\ \hline
\% \textbf{\texttt{CELL}} contradicts& \textbf{75.2} & \textbf{75.2} & \textbf{80.2}& \textbf{81.0} & \textbf{85.0} & \textbf{84.2} & \textbf{83.8} \\ \hline
\% Baseline contradicts & 13.2 & 11.8 & 12.4 & 14.4 & 11.8 & 11.8 & 15.0 \\ \hline

\end{tabular}
\end{table}

\subsection{Preference Comparisons}
\label{ss:contrast_comparisons}
\eat{We investigate the quality of the contrast (i.e., the response to the perturbed prompt) generated by \textbf{\texttt{CELL}} compared to \textbf{\texttt{Baseline}}. The template prompt used to generate the less preferable response is: \texttt{``Answer the following prompt in one sentence, less than 20 words, and to the point: Give a less preferable response than \{\} to the prompt: \{\}.''} The two \{\}'s contain the input response and prompt, respectively. This template was finalized after several variations and manual inspections. It is crucial to recall that no baselines exist in previous literature; hence we pursue this baseline as a natural comparison.}

\begin{table}[t]
\small
\centering
\caption{Average preference scores comparing \texttt{Llama}, \textbf{\texttt{CELL}}, and \textbf{\texttt{Baseline}} responses. Positive numbers for \texttt{Llama} vs \textbf{\texttt{CELL}} represent a higher preference for responses from \texttt{Llama} than \textbf{\texttt{CELL}} (similarly for Baseline vs \textbf{\texttt{m-CELL}}). Higher \#s (i.e. lower preference for \textbf{\texttt{CELL}}) indicate \textbf{\texttt{CELL}} is better, which is the desired effect. \textbf{\texttt{m-CL}} denotes \textbf{\texttt{m-CELL}} which shows similar trends. \eat{The positive numbers overall signifies that the initial \texttt{Llama} responses and Baseline responses were found to be preferable to \textbf{\texttt{(m-)CELL}} responses, which is the desired effect of the algorithms.}}
\label{tab:llama_preference_baseline}
\footnotesize
\setlength{\tabcolsep}{2.5pt}
\begin{tabular}{|c|c|c|c|c|c|c|c|c|}
\multicolumn{9}{c}{\textbf{Scoring function: Preference}} \\ \hline
\multirow{3}{*}{\rotatebox[origin=c]{90}{split\_k}}& \multicolumn{4}{|c|}{\textbf{Llama vs \texttt{(m-)CELL}}} & \multicolumn{4}{|c|}{\textbf{Baseline vs \texttt{(m-)CELL}}} \\ \cline{2-9}
 & \multicolumn{2}{|c|}{\textbf{Llama3-8b}} & \multicolumn{2}{|c|}{\textbf{Llama3-70b}}& \multicolumn{2}{|c|}{\textbf{Llama3-8b}} & \multicolumn{2}{|c|}{\textbf{Llama3-70b}} \\ \cline{2-9}
& \textbf{\texttt{m-CL}} & \textbf{\texttt{CELL}} & \textbf{\texttt{m-CL}} & \textbf{\texttt{CELL}} & \textbf{\texttt{m-CL}} & \textbf{\texttt{CELL}} & \textbf{\texttt{m-CL}} & \textbf{\texttt{CELL}} \\ \hline
1&0.31&0.32&0.32&0.32&0.11&0.12&0.07&0.07\\\hline
2&0.32&0.35&0.32&0.34&0.15&0.19&0.08&0.13\\\hline
3&0.31&0.35&0.32&0.35&0.15&0.2&0.09&0.14\\\hline
\end{tabular}
\end{table}

We here investigate the quality of contrasts generated by \textbf{\texttt{CELL}} compared to \textbf{\texttt{Baseline}}. Results are shown in Table \ref{tab:llama_preference_baseline} applying \texttt{Llama} LLMs to prompts from the MIC dataset. Each entry dictates which response is preferred for a given input prompt as measured by the scoring function \textbf{preference} defined in Section \ref{ss:scoring_functions}. Table \ref{tab:llama_preference_baseline} (left) compares responses of the corresponding \texttt{Llama} LLMs to the contrastive responses output by \textbf{\texttt{(m-)CELL}} and the Table \ref{tab:llama_preference_baseline} (right) compares \textbf{\texttt{Baseline}} to \textbf{\texttt{(m-)CELL}}. Each row is for a different value of \textbf{\texttt{CELL}} parameter \texttt{split\_k}.

We observe that \textbf{\texttt{CELL}} and \textbf{\texttt{m-CELL}} produce similar results across different values of \texttt{split\_k}, likely due to short prompt lengths in MIC. Importantly, positive numbers mean that \textbf{\texttt{CELL}} responses were found to be less preferable to the initial \texttt{Llama} and \textbf{\texttt{Baseline}} responses. Corresponding experiments using the \textbf{contradiction} scoring function can be found in the Appendix Section \ref{s:additional_quant_experiments}.

\begin{table}[t] 
\small
\centering
\caption{Average edit distances and flip rates comparing \textbf{\texttt{CELL}} vs \textbf{\texttt{m-CELL}}. Smaller edit rates, larger flip rates are better.}
\label{tab:edit_flip}
\footnotesize
\setlength{\tabcolsep}{2.5pt}
\begin{tabular}{|c|c|c|c|c|c|c|c|c|}
\multicolumn{9}{c}{\textbf{Scoring function: Preference}} \\ \hline
\multirow{3}{*}{\rotatebox[origin=c]{90}{split\_k}}& \multicolumn{4}{|c|}{\textbf{Average Edit Distance}} & \multicolumn{4}{|c|}{\textbf{Average Flip Rate}} \\ \cline{2-9}
 & \multicolumn{2}{|c|}{\textbf{Llama3-8b}} & \multicolumn{2}{|c|}{\textbf{Llama3-70b}}& \multicolumn{2}{|c|}{\textbf{Llama3-8b}} & \multicolumn{2}{|c|}{\textbf{Llama3-70b}} \\ \cline{2-9}
& \textbf{\texttt{m-CL}} & \textbf{\texttt{CELL}} & \textbf{\texttt{m-CL}} & \textbf{\texttt{CELL}} & \textbf{\texttt{m-CL}} & \textbf{\texttt{CELL}} & \textbf{\texttt{m-CL}} & \textbf{\texttt{CELL}}\\ \hline
1&0.12&0.16&0.13&0.17&0.85&0.89&0.88&0.88\\\hline
2&0.19&0.21&0.17&0.22&0.9&0.95&0.89&0.95\\\hline
3&0.25&0.29&0.24&0.29&0.88&0.95&0.86&0.93\\\hline
\multicolumn{9}{c}{} \\ 
\multicolumn{9}{c}{\textbf{Scoring function: Contradiction}} \\ \hline
\multirow{3}{*}{\rotatebox[origin=c]{90}{split\_k}}& \multicolumn{4}{|c|}{\textbf{Average Edit Distance}} & \multicolumn{4}{|c|}{\textbf{Average Flip Rate}} \\ \cline{2-9}
 & \multicolumn{2}{|c|}{\textbf{Llama3-8b}} & \multicolumn{2}{|c|}{\textbf{Llama3-70b}}& \multicolumn{2}{|c|}{\textbf{Llama3-8b}} & \multicolumn{2}{|c|}{\textbf{Llama3-70b}} \\ \cline{2-9}
& \textbf{\texttt{m-CL}} & \textbf{\texttt{CELL}} & \textbf{\texttt{m-CL}} & \textbf{\texttt{CELL}} & \textbf{\texttt{m-CL}} & \textbf{\texttt{CELL}} & \textbf{\texttt{m-CL}} & \textbf{\texttt{CELL}}\\ \hline
1&0.17&0.19&0.16&0.19&0.56&0.58&0.62&0.66\\\hline
2&0.25&0.26&0.24&0.25&0.59&0.71&0.66&0.79\\\hline
3&0.34&0.32&0.31&0.32&0.54&0.68&0.63&0.78\\\hline
\eat{
\multicolumn{9}{c}{} \\ \hline
\multirow{3}{*}{\rotatebox[origin=c]{90}{split\_k}}& \multicolumn{4}{|c|}{\textbf{Number of Model Calls}} & \multicolumn{4}{|c|}{\textbf{Time (seconds)}} \\ \cline{2-9}
 & \multicolumn{2}{|c|}{\textbf{Llama3-8b}} & \multicolumn{2}{|c|}{\textbf{Llama3-70b}}& \multicolumn{2}{|c|}{\textbf{Llama3-8b}} & \multicolumn{2}{|c|}{\textbf{Llama3-70b}} \\ \cline{2-9}
& \textbf{\texttt{m-CL}} & \textbf{\texttt{CELL}} & \textbf{\texttt{m-CL}} & \textbf{\texttt{CELL}} & \textbf{\texttt{m-CL}} & \textbf{\texttt{CELL}} & \textbf{\texttt{m-CL}} & \textbf{\texttt{CELL}}\\ \hline
1&30.9&28.3&28.8&28.7&35.5&32.9&61.8&63.3\\\hline
2&15.2&20.2&14.1&19.7&17.8&23.7&29.7&41.6\\\hline
3&10.2&16.3&9.7&15.5&12.2&19.3&19.9&32.7\\\hline
}\end{tabular}
    \vspace{-0.25cm}
\end{table}

\subsection{Contrastive Explanation Properties}
We first evaluate \textbf{\texttt{CELL}} perturbed prompts across 2 main properties considered in works on contrastive explanations for classifiers \citep{cat, mice}: flip and edit rates. The flip rate measures the percentage of times \textbf{\texttt{CELL}} finds a contrastive explanation. Edit distances compute a word-level Levenstein distance between the input and contrastive prompt. This is the minimum number of changes (additions, deletions, etc.) to get from one prompt to the other, and we normalize by the number of words in the input prompt. \eat{Content preservation quantifies how much content is preserved in the contrastive prompt from the input prompt. Following previous works above, we compute the cosine similarity between prompt embeddings obtained from a \texttt{bert-base-uncased} model.} Table \ref{tab:edit_flip} shows edit distances and flip rates using the MIC data with both \textbf{preference} and \textbf{contradiction} scoring functions. The results show that \textbf{\texttt{CELL}} finds more contrasts with only slightly larger distance using an intelligent, rather than myopic, search. Edit distances are comparable to previous literature for explaining classifiers \citep{cat,mice}. Flip rates are lower than in those works\eat{ where the flip rate is typically $\ge$ 0.95}, but this reflects the difficulty in explaining LLMs versus classifiers for which many of the methods are not black-box and have access to gradients for selecting important words. \eat{Content preservation was found to be $\ge 0.99$ across all models and scoring functions which is significantly higher than most results seen in \citet{cat} and  \citet{mice}, likely due to the better and more flexible infilling models used here.}

\eat{Table \ref{tab:edit_flip} (bottom) shows the average number of model calls made both by algorithms. We observe here that \textbf{\texttt{CELL}} is not always more efficient; \textbf{\texttt{m-CELL}} requires less model calls and hence is faster than \textbf{\texttt{CELL}} when \texttt{split\_k=3} because of the reduced search space, but \textbf{\texttt{CELL}} is more efficient than \textbf{\texttt{m-CELL}} when \texttt{split\_k=1} due to the larger search space. Interestingly, the number of model calls is similar for both \texttt{Llama} LLMs. Putting these trends together with those from Table \ref{tab:llama_preference_baseline} suggests the slightly better quality can sometimes be obtained with the higher \texttt{split\_k=3} which is also more efficient here using \textbf{\texttt{m-CELL}} rather than \textbf{\texttt{CELL}}. Corresponding experiments using the \textbf{contradiction} scoring function are in the Appendix Section \ref{s:additional_quant_experiments}. We also hold to the Appendix Section \ref{s:longer_documents} further experiments demonstrating \textbf{\texttt{CELL}} on longer documents from the Extreme Summarization (XSum) dataset \citep{xsum-emnlp}.
}

\begin{figure}[t]
    \centering
    \includegraphics[width=0.5\textwidth]{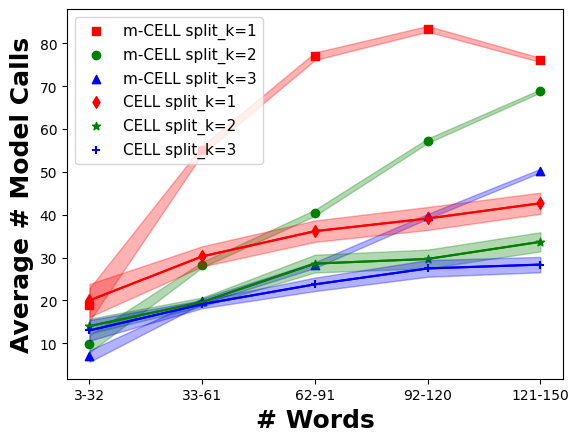}
    \vspace{-0.5cm}
      \caption{\small{Average \# model calls for \textbf{\texttt{CELL}} and \textbf{\texttt{m-CELL}} applied to text summarization. Shaded regions denote standard error.}
    }
    \label{fig:efficiency_xsum}
    \vspace{-0.5cm}
\end{figure}

We consider the number of model calls by \textbf{\texttt{CELL}} and \textbf{\texttt{m-CELL}} on longer documents from the Extreme Summarization (XSum) dataset \citep{xsum-emnlp} in a text summarization task, illustrated in Figure \ref{fig:efficiency_xsum}. These explanations use the \textbf{BLEU\_SUMM} scoring function defined in Section \ref{ss:scoring_functions}. Note that other recent explainability works for LLMs \citep{mexgen, textgenshap}, albeit attribution methods, do not report on such efficiency statistics which are known to be important in practice \citep{macem}. The figure illustrates how the number of model calls vary across different length documents as well as different values of parameter \texttt{split\_k}. As \textbf{\texttt{m-CELL}} does not scale well for large documents, a 2 minute limit for explanations is set. \textbf{\texttt{m-CELL}} is competitive on shorter documents (as also noted by smaller edit distances in Table \ref{tab:edit_flip}), but clearly is not as competitive on longer documents. Interestingly, the number of \textbf{\texttt{CELL}} model calls plateaus, illustrating the effectiveness of the budgeted strategy. \eat{We also observe that setting  \texttt{split\_k} $>1$ shows improvement on large texts (due to a reduced search space), but statistically, we do not see reason to go beyond \texttt{split\_k} $= 2$ here.} 

\subsection{Application: Explaining Conversations}

\begin{figure*}[t]
    \centering
    \includegraphics[width=.9\textwidth]{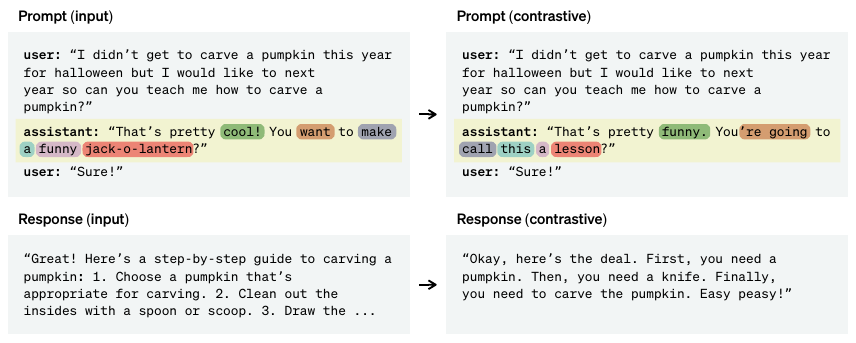}%
    \vspace{0.3em}
    \hrule\hrule
    \vspace{0.2em}
    \includegraphics[width=.9\textwidth]{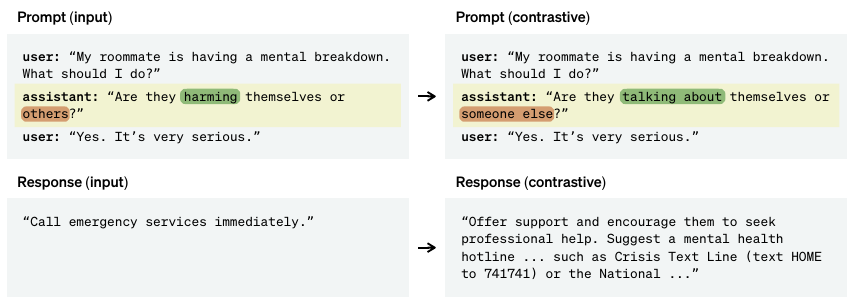}
    \caption{Example of explaining conversations. Colors match between what is changed between input and contrastive prompts (restricted to the assistant).}
    \label{fig:deg}
\vspace{-0.25cm}
\end{figure*}

Modern LLMs are able to maintain coherent conversations with human users over many turns. However, current models still suffer from various issues, especially in long-form dialogue \citep{setlur2022you,shaikh2023grounding}. 
Improving a model’s conversational capability requires fine-grained identification of their weak points and subsequently obtaining training data to fill these gaps.

In this section, we illustrate how our contrastive explanations method can be used to help understand why models generate specific responses subject to the conversational context. 
We define a class of metrics, adopted from definitions of conversational quality from various sources \citep{grice1975logic, higashinaka2019improving, galetzka2023neural, miehling2024language}, spanning six submaxims (quantity, quality, relevance, manner, benevolence, transparency) to evaluate conversational turns. Each metric takes as input a \emph{context} (history of turns) and a \emph{response} (the most recent turn) and generates a score on a particular submaxim dimenion using an LLM-based labeling procedure. 

To label turns, we create scoring rubrics that reflect the submaxims according to definitions and requirements taken from \citet{miehling2024language}. These scoring rubrics are constructed by first describing (in natural language) the requirement of each submaxim, and then including in-context examples to aid the model with the labeling task. The turn to evaluate is then appended to the prompt. We use \path{mistralai/Mixtral-8x7B-Instruct-v0.1} to generate the labels. Additionally, we query the model multiple times, and average the resulting scores, to obtain a more robust label. Sample prompts for helpfulness, harm, and informativeness can be found in the Appendix Section \ref{app:degradation}.

These metrics can thus be viewed as LLM-as-a-Judge metrics (similar to how we use Prometheseus2 in Section \ref{ss:llm_as_a_judge}).  {\bf \texttt{CELL}} is leveraged to explain each label by minimally modifying the previous assistant turn(s) until the threshold is passed. 
Fig. \ref{fig:deg} illustrates two example conversations. The top example presents a helpful assistant response (input) that gives clear directions, which, after modifying the assistant's turn in the prompt to be less serious and relevant, results in a significantly less helpful response (contrastive) that gives insufficient directions. The bottom example contains an initial assistant response that recognizes the emergency of the stated situation. Modifying the previous assistant response results in a less urgent response (contrastive), illustrating that the cause of the original urgency was the statement that the roommate was harming themself. Such explanations could be used to generate training data to improve conversational agents. One might want to generate data where one dimension is modified and the others remain fixed. For example, by explaining the top example in Figure \ref{fig:deg} over other submaxims, it could potentially be used as an example of not being helpful while maintaining other submaxims. Additional examples of explaining conversations can be found in the Appendix Section \ref{s:additional_qualitiative_examples}.

\section{Conclusion}
This paper proposes novel contrastive explanations for large language models. Novel insight into what a contrast should mean regarding LLMs led us to propose two algorithms for generating contrastive explanations: a myopic method that is effective for explaining responses to small prompts and a novel search strategy that takes into account a model query budget. 
Our novel use case explicitly provides actionable explanations. As suggested, one might want to fix certain submaxims while explaining conversations in terms of other dimensions. This suggests future algorithmic work, where we would like to adapt \textbf{\texttt{CELL}} so that the search explicitly moves in such directions.

\clearpage
\eat{
\section{Impact Statement}
Contrastive explanations have the potential to debug LLMs by discovering prompts that lead to toxic or other undesireable responses, or, as noted in the conclusion, could help develop training data to improve conversational AI agents. As with many technologies, contrastive explanations also have potential consequences. Our explanations are limited in that they cannot be prevented from providing explanations that are offensive or harmful. If an LLM has the potential to be harmful, it is conceivable that our explanations could find a harmful prompt that generates a harmful response that is less preferable than an input response. The method is designed to explain an LLM and is subject to the capabilities of the LLM it is explaining. This is a limitation of all contrastive explanations, whether for images, tabular data, or as noted here, for Large Language Models. Another potential impact is on reducing the cost of such explanations when considering large documents, due to the intelligent search method; however, it remains limited by the query budget. For example, Question/Answering tasks typically contain the answer as a short substring of a large context. As the search involves randomly sampling where in the context to look for the answer, there is a strictly positive likelihood that such answers are never inspected in the context. Our other method that is not limited by a query budget is still limited by computational resources as in the worst case, it can query the LLM $n!$ times. 
}
\bibliographystyle{plainnat}
\bibliography{main-arXiv}

\appendix

\section{Pseudocodes}
This section contains several algorithms described in the paper. Algorithm \ref{algo:cem_llm_greedy} is the pseudocode for \textbf{\texttt{m-CELL}}. Algorithms \ref{algo:num_centers}, \ref{algo:generate_centers}, and \ref{algo:sample_centers} are the helper functions to \textbf{\texttt{CELL}}. One other assumed function for \textbf{\texttt{SAMPLE\_CENTERS}} is a function \textbf{\texttt{SAMPLE}} that outputs $m$ random entries from any set $\mathcal{S}$. 

\begin{algorithm}[h]
\caption{{\bf \texttt{m-CELL}}}
  \label{algo:cem_llm_greedy} 
\begin{algorithmic}

\STATE \textbf{Input:} $LLM(\cdot)$, infilling model $I(\cdot)$, scoring function $g(\cdots)$, threshold $\delta$, prompt $x_0$

\STATE $Z \leftarrow  \mbox{split\_prompt}(x), \quad n_e \leftarrow |Z|$

\STATE  $\mathcal{J} \leftarrow \{1,\ldots,n_e\}$ \# unmasked substring indices

\STATE  $x_c \leftarrow x$

\STATE  \# Loop to select substrings to mask

\FOR{$i=1$ to $n_e$}
    \STATE$y_c \leftarrow LLM(x_c)$
    \FOR{$j\in\mathcal{J}$}
        \STATE $x_j \leftarrow I(\mbox{mask}(x_c, Z, j))$ 
        \STATE $y_j \leftarrow LLM(x_j)$ 
        \STATE $z_j \leftarrow g(x_c, y_c, y_j)$
    \ENDFOR
    \STATE $j^* \leftarrow \argmax_{j\in\mathcal{J}}{z_j}$

    \IF{$z_{j^*} \ge \delta:$}{
        \STATE \textbf{Output:} $(x_0, LLM(x_0), x_{j^*}, y_{j^*})$}
    \ELSE
        \STATE $\mathcal{J} \leftarrow \mathcal{J} / j^*$
    \ENDIF
    \STATE $x_c \leftarrow x_{j^*}$
\ENDFOR
\STATE PRINT('NO SOLUTION FOUND')
\end{algorithmic}
\end{algorithm}

\begin{algorithm}[h]
\caption{{\bf \texttt{NUM\_SEEDS}}}
  \label{algo:num_centers} 
\begin{algorithmic}
\STATE \textbf{Input:} iteration number $t$, Budget $B$
\STATE $q \leftarrow \mbox{floor}(B/\log{(B)})$
\IF{$(t+1)\cdot  2^t \leq q$}
    \STATE $m = 2^{t+1}$ 
\ELSE
    \STATE $m=2^t$
\ENDIF
\STATE \textbf{Ouput: } $m$
\end{algorithmic}
\end{algorithm}

\begin{algorithm}[h]
\caption{{\bf \texttt{GENERATE\_SEEDS}}}
  \label{algo:generate_centers} 
\begin{algorithmic}
\STATE \textbf{Input:} number of seeds to generate $m$, current list of triples of (perturbed prompt, unmasked substring indices list, score) $X_F$, list of current unmasked substring indices $\mathcal{J}$, percentage seeds from ratio $\alpha$, prompt $x_0$, list of split prompt tokens $Z$
\STATE $m_1 \leftarrow \min{(\alpha\cdot m, |X_F|)}$
\STATE $m_2 \leftarrow \min{(m-m_1, |\mathcal{J}|)}$
\STATE $\mathcal{I}_1 \leftarrow \mbox{SAMPLE}(X_F, m_1)$
\STATE $\mathcal{I}_2 \leftarrow \mbox{SAMPLE}(\mathcal{J}, m_2)$
\STATE $X_c \leftarrow \{\quad\}$ \# list of current seeds
\STATE \# Perturb $m_1$ perturbed prompts
\FOR{$(x_s, J_s, f_s)\in\mathcal{I}_1$}
    \STATE $j \leftarrow SAMPLE(J_s, 1)$
    \STATE $J_s \leftarrow J_s / \{j\}$
    \STATE $X_c \leftarrow X_c \cup \{(I(\mbox{mask}(x_s, Z, j), J_s)\}$
\ENDFOR
\STATE \# Perturb $m_2$ tokens from initial prompt
\FOR{$j\in\mathcal{I}_2$}
    \STATE $X_c \leftarrow X_c \cup \{(I(\mbox{mask}(x, Z, j), \mathcal{J} / \{j\})\}$
\ENDFOR
\STATE \textbf{Output: }$X_c$
\end{algorithmic}
\end{algorithm}

\begin{algorithm}[h]
\caption{{\bf \texttt{SAMPLE\_SEEDS}}}
  \label{algo:sample_centers} 
\begin{algorithmic}
\STATE \textbf{Input:} list of prompt seeds $X_c$, list of split prompt tokens $Z$, \# samples per seed $n_s$, Infiller $I(\cdot)$
\STATE \# Sample around all prompt seeds
\STATE $X_p \leftarrow \{\quad\}$
\FOR{$(x_s, J_s)\in X_c$}
    \FOR{$j=1$ to $n_s$}
        \STATE $j \leftarrow SAMPLE(J_s, 1)$
        \STATE $J_t \leftarrow J_s / \{j\}$
        \STATE $X_p \leftarrow X_p \cup \{(I(\mbox{mask}(x_s, Z, j), J_t)\}$
    \ENDFOR
\ENDFOR
\STATE $X_p \leftarrow X_P \cup X_c$
\STATE \textbf{Output: }$X_p$
\end{algorithmic}
\end{algorithm}

\section{Additional Related Work: General Local XAI Methods}
\label{s:additional_related_work}
\citet{xainlp_survey} considered explainability for natural language processing, primarily for classification tasks where local explanations were provided, among which our focus is on post-hoc methods that explain a fixed model's prediction. One large group of explainability methods are feature based where the explanation outputs some form of feature importance (i.e., ranking, relevance, etc.) of the words in text \citep{dknn_text,dknn,inputreduction,patternattribution_text,lime,melis_jaakkola_2017}. Other types of local post-hoc explanations include exemplar based \citep{proto,infl,l2c} that output similar instances to the input.

\section{Additional quantitative experiments and statistics}
\label{s:additional_quant_experiments}
Additional experimental results are given in this section. Table \ref{tab:llama_preference_baseline_appendix} corresponds to the results in the Preference Comparisons subsection albeit with the contradiction metric. Standard errors for experiments in the paper are given by Tables \ref{tab:llama_preference_baseline_sterror_appendix} and \ref{tab:llama_efficiency_sterror}. 

\section{Prometheus2 Template}
\label{app:prometheseus2}
\begin{figure*}
    \begin{center}
\includegraphics[width=0.75\textwidth]{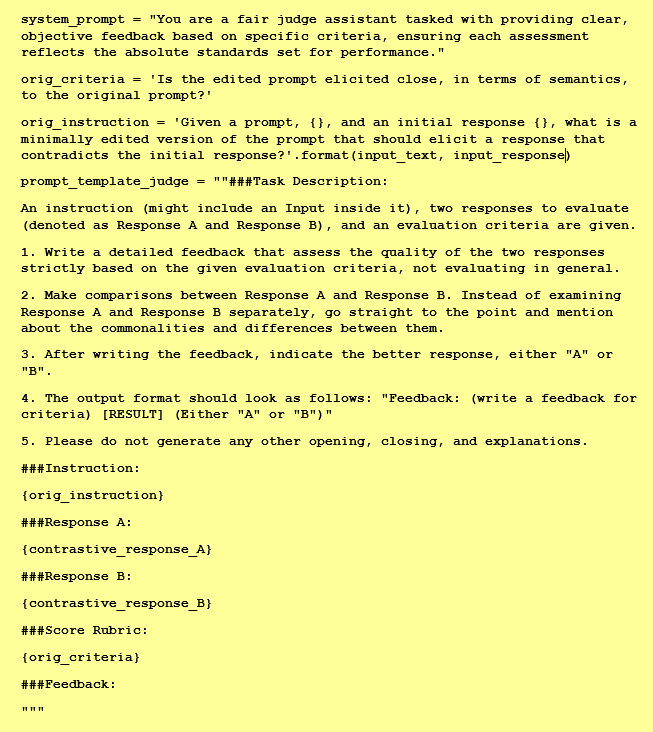}
\caption{Prompt used for Prometheus2.}
\label{fig:prometheus2_prompt}
\end{center}
\end{figure*}

Prometheus2 is used an an LLM-as-a-Judge in Section \ref{ss:llm_as_a_judge}. This requires a particular template which is displayed in Figure \ref{fig:prometheus2_prompt}. Four temporary strings are replaced in the template called \path{prompt_template_judge}: \path{orig_instruction}, \path{contrastive_response_A}, \path{contrastive_response_B}, and \path{orig_criteria}. For the \textbf{preference} scoring function, the \path{orig_instruction} is given in the figure, where the two blank entries are filled with the input prompt and corresponding response by the LLM being explained. The two responses being compared by the judge replace \path{contrastive_response_A} and \path{contrastive_response_B}. Lastly, \path{orig_criteria} is also given in the figure. The \path{system_prompt} at the top of the figure is appended to the final \path{prompt_template_judge} which is then passed to the judge.

\section{Additional qualitative examples}
\label{s:additional_qualitiative_examples}
\begin{figure*}[t]
    \centering
    \includegraphics[width=\textwidth]{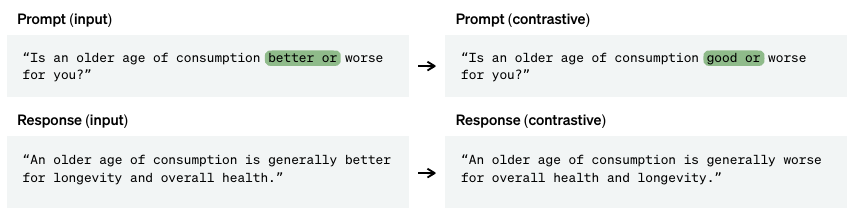}%
    \vspace{0.3em}
    \hrule\hrule
    \vspace{0.2em}
    \includegraphics[width=\textwidth]{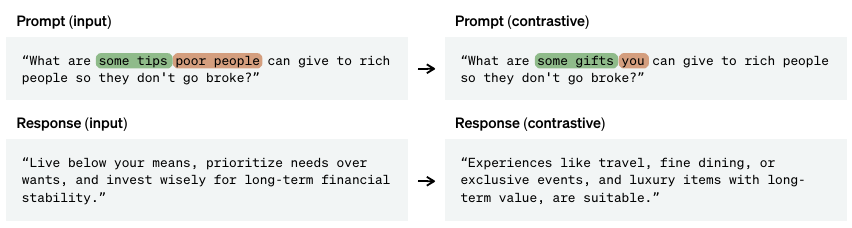}
    \caption{Additional contrastive explanations for natural language generation by \texttt{meta-llama/Llama-3.1-8B-Instruct}. Colors match between what is changed between input prompt and contrastive prompt. These explanations explain that the Input Prompt generated the input response because if the modifications were made to the input Prompt, the new contrastive prompt would generate the contrastive response instead, which contradicts the input response (as per NLI model \texttt{cross-encoder/nli-deberta-v3-base}). Prompts taken from the Moral Integrity Corpus \citep{micdata}.}
    \label{fig:app_exs}
\end{figure*}

\begin{figure*}[t]
    \centering
    \includegraphics[width=\textwidth]{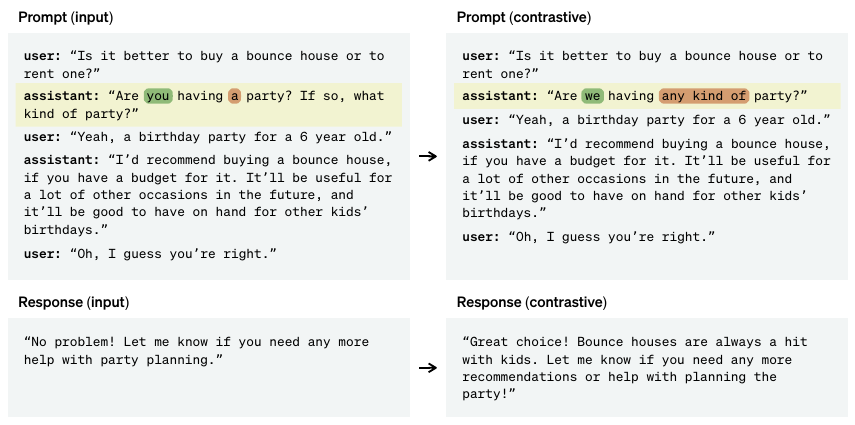}%
    \vspace{0.3em}
    \hrule\hrule
    \vspace{0.2em}
    \includegraphics[width=\textwidth]{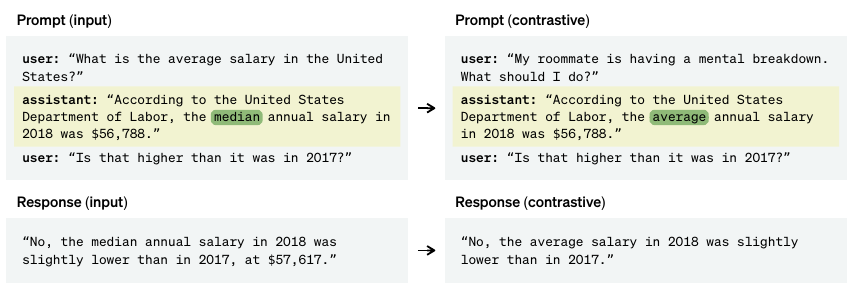}
    \caption{Additional examples of explanations for conversational degradation. The top example presents a simple assistant response (in input response) but after modifying the assistant's first of two turns to make the assistant seem more \emph{excited} and part of the event, the resulting response is more informative. The bottom example illustrates an example where changing a single word can alter the response to decrease helpfulness (since a different question is answered). This is often possible on simple question/answer type scenarios.}
    \label{fig:app_exs_conv}
\end{figure*}

Two additional examples on natural language generation from the MIC data can be found in Fig. \ref{fig:app_exs}. Two additional examples explaining conversations can be found in Fig. \ref{fig:app_exs_conv}. 

\begin{table}[h] 
\small
\centering
\caption{Average preference scores comparing Llama, \textbf{\texttt{CELL}} responses, and \textbf{\texttt{Baseline}} contrastive response. Positive numbers for Llama vs \textbf{\texttt{CELL}} represent a higher preference for responses from Llama than \textbf{\texttt{CELL}} (similarly for \textbf{\texttt{Baseline}} vs \textbf{\texttt{CELL}}). Higher \#s (i.e. lower preference for \textbf{\texttt{CELL}}) indicate \texttt{CELL} is better. K refers to the \texttt{split\_k} parameter. \textbf{\texttt{m-CL}} denotes \textbf{\texttt{m-CELL}}. Contrastive explanations here were generated using a contradiction metric. }
\label{tab:llama_preference_baseline_appendix}
\footnotesize
\setlength{\tabcolsep}{2.5pt}
\begin{tabular}{|c|c|c|c|c|c|c|c|c|}
\multicolumn{9}{c}{\textbf{Metric: Contradiction}} \\ \hline
\multirow{3}{*}{\rotatebox[origin=c]{90}{split\_k}}& \multicolumn{4}{|c|}{\textbf{Llama vs \texttt{(m-)CELL}}} & \multicolumn{4}{|c|}{\textbf{Baseline vs \texttt{(m-)CELL}}} \\ \cline{2-9}
 & \multicolumn{2}{|c|}{\textbf{Llama3-8b}} & \multicolumn{2}{|c|}{\textbf{Llama3-70b}}& \multicolumn{2}{|c|}{\textbf{Llama3-8b}} & \multicolumn{2}{|c|}{\textbf{Llama3-70b}} \\ \cline{2-9}
& \textbf{\texttt{m-CL}} & \textbf{\texttt{CELL}} & \textbf{\texttt{m-CL}} & \textbf{\texttt{CELL}} & \textbf{\texttt{m-CL}} & \textbf{\texttt{CELL}} & \textbf{\texttt{m-CL}} & \textbf{\texttt{CELL}} \\ \hline
1&0.19&0.2&0.2&0.21&0.03&0.04&0.04&0.05\\\hline
2&0.22&0.23&0.22&0.22&0.06&0.06&0.06&0.1\\\hline
3&0.22&0.22&0.22&0.22&0.08&0.08&0.08&0.09\\\hline
\end{tabular}
\end{table}

\begin{table}[t] 
\small
\centering
\caption{Standard errors of average preference scores comparing Llama, \textbf{\texttt{CELL}} responses, and \textbf{\texttt{Baseline}} contrastive response, both using preference and contradiction as the metric. Results generated from 500 prompts taken from the Moral Integrity Corpus (test split). \textbf{\texttt{m-CL}} denotes \textbf{\texttt{m-CELL}}.}
\label{tab:llama_preference_baseline_sterror_appendix}
\footnotesize
\setlength{\tabcolsep}{2.5pt}
\begin{tabular}{|c|c|c|c|c|c|c|c|c|}
\multicolumn{9}{c}{\textbf{Metric: Preference}} \\
\multicolumn{9}{c}{\textbf{Standard Errors}} \\ \hline
\multirow{3}{*}{\rotatebox[origin=c]{90}{split\_k}}& \multicolumn{4}{|c|}{\textbf{Llama vs \texttt{(m-)CELL}}} & \multicolumn{4}{|c|}{\textbf{Baseline vs \texttt{(m-)CELL}}} \\ \cline{2-9}
 & \multicolumn{2}{|c|}{\textbf{Llama3-8b}} & \multicolumn{2}{|c|}{\textbf{Llama3-70b}}& \multicolumn{2}{|c|}{\textbf{Llama3-8b}} & \multicolumn{2}{|c|}{\textbf{Llama3-70b}} \\ \cline{2-9}
& \textbf{\texttt{m-CL}} & \textbf{\texttt{CELL}} & \textbf{\texttt{m-CL}} & \textbf{\texttt{CELL}} & \textbf{\texttt{m-CL}} & \textbf{\texttt{CELL}} & \textbf{\texttt{m-CL}} & \textbf{\texttt{CELL}} \\ \hline
1&0.01&0.01&0.01&0.01&0.03&0.03&0.03&0.03\\\hline
2&0.01&0.01&0.01&0.01&0.03&0.03&0.03&0.03\\\hline
3&0.01&0.01&0.01&0.01&0.02&0.02&0.03&0.03\\\hline
\multicolumn{9}{c}{} \\ 
\multicolumn{9}{c}{\textbf{Metric: Contradiction}} \\
\multicolumn{9}{c}{\textbf{Standard Errors}} \\ \hline
\multirow{3}{*}{\rotatebox[origin=c]{90}{split\_k}}& \multicolumn{4}{|c|}{\textbf{Llama vs \texttt{(m-)CELL}}} & \multicolumn{4}{|c|}{\textbf{Baseline vs \texttt{(m-)CELL}}} \\ \cline{2-9}
 & \multicolumn{2}{|c|}{\textbf{Llama3-8b}} & \multicolumn{2}{|c|}{\textbf{Llama3-70b}}& \multicolumn{2}{|c|}{\textbf{Llama3-8b}} & \multicolumn{2}{|c|}{\textbf{Llama3-70b}} \\ \cline{2-9}
& \textbf{\texttt{m-CL}} & \textbf{\texttt{CELL}} & \textbf{\texttt{m-CL}} & \textbf{\texttt{CELL}} & \textbf{\texttt{m-CL}} & \textbf{\texttt{CELL}} & \textbf{\texttt{m-CL}} & \textbf{\texttt{CELL}} \\ \hline
1&0.01&0.01&0.01&0.01&0.03&0.03&0.02&0.02\\\hline
2&0.01&0.01&0.01&0.01&0.03&0.03&0.02&0.02\\\hline
3&0.01&0.01&0.01&0.01&0.03&0.03&0.03&0.02\\\hline
\end{tabular}
\end{table}

\begin{table}[t] 
\vspace{-4mm}
\small
\centering
\caption{Standard errors of edit distances and flip rates comparing \textbf{\texttt{CELL}} vs \textbf{\texttt{m-CELL}} on prompts from the Moral Integrity Corpus, both using preference and contradiction metrics. \textbf{\texttt{m-CL}} denotes \textbf{\texttt{m-CELL}}.}
\label{tab:llama_efficiency_sterror}
\footnotesize
\setlength{\tabcolsep}{2.5pt}
\begin{tabular}{|c|c|c|c|c|c|c|c|c|}
\multicolumn{9}{c}{\textbf{Metric: Preference}} \\ \hline
\multirow{3}{*}{\rotatebox[origin=c]{90}{split\_k}}& \multicolumn{4}{|c|}{\textbf{Std Error  Edit Distance}} & \multicolumn{4}{|c|}{\textbf{Std Error Flip Rate}} \\ \cline{2-9}
 & \multicolumn{2}{|c|}{\textbf{Llama3-8b}} & \multicolumn{2}{|c|}{\textbf{Llama3-70b}}& \multicolumn{2}{|c|}{\textbf{Llama3-8b}} & \multicolumn{2}{|c|}{\textbf{Llama3-70b}} \\ \cline{2-9}
& \textbf{\texttt{m-CL}} & \textbf{\texttt{CELL}} & \textbf{\texttt{m-CL}} & \textbf{\texttt{CELL}} & \textbf{\texttt{m-CL}} & \textbf{\texttt{CELL}} & \textbf{\texttt{m-CL}} & \textbf{\texttt{CELL}}\\ \hline
1&0.01&0.01&0.01&0.01&0.02&0.01&0.01&0.01\\\hline
2&0.01&0.01&0.01&0.01&0.01&0.01&0.01&0.01\\\hline
3&0.01&0.01&0.01&0.01&0.01&0.01&0.02&0.01\\\hline
\multicolumn{9}{c}{} \\ 
\multicolumn{9}{c}{\textbf{Metric: Contradiction}} \\ \hline
\multirow{3}{*}{\rotatebox[origin=c]{90}{split\_k}}& \multicolumn{4}{|c|}{\textbf{Std Error  Edit Distance}} & \multicolumn{4}{|c|}{\textbf{Std Error Flip Rate}} \\ \cline{2-9}
 & \multicolumn{2}{|c|}{\textbf{Llama3-8b}} & \multicolumn{2}{|c|}{\textbf{Llama3-70b}}& \multicolumn{2}{|c|}{\textbf{Llama3-8b}} & \multicolumn{2}{|c|}{\textbf{Llama3-70b}} \\ \cline{2-9}
& \textbf{\texttt{m-CL}} & \textbf{\texttt{CELL}} & \textbf{\texttt{m-CL}} & \textbf{\texttt{CELL}} & \textbf{\texttt{m-CL}} & \textbf{\texttt{CELL}} & \textbf{\texttt{m-CL}} & \textbf{\texttt{CELL}}\\ \hline
1&0.01&0.01&0.01&0.01&0.02&0.02&0.02&0.02\\\hline
2&0.01&0.01&0.01&0.01&0.01&0.02&0.02&0.02\\\hline
3&0.01&0.01&0.01&0.01&0.02&0.02&0.02&0.02\\\hline

\end{tabular}
\end{table}

\section{Sample prompts for labeling conversations}
\label{app:degradation}

\begin{figure*}
\begin{center}
\includegraphics[width=\textwidth]{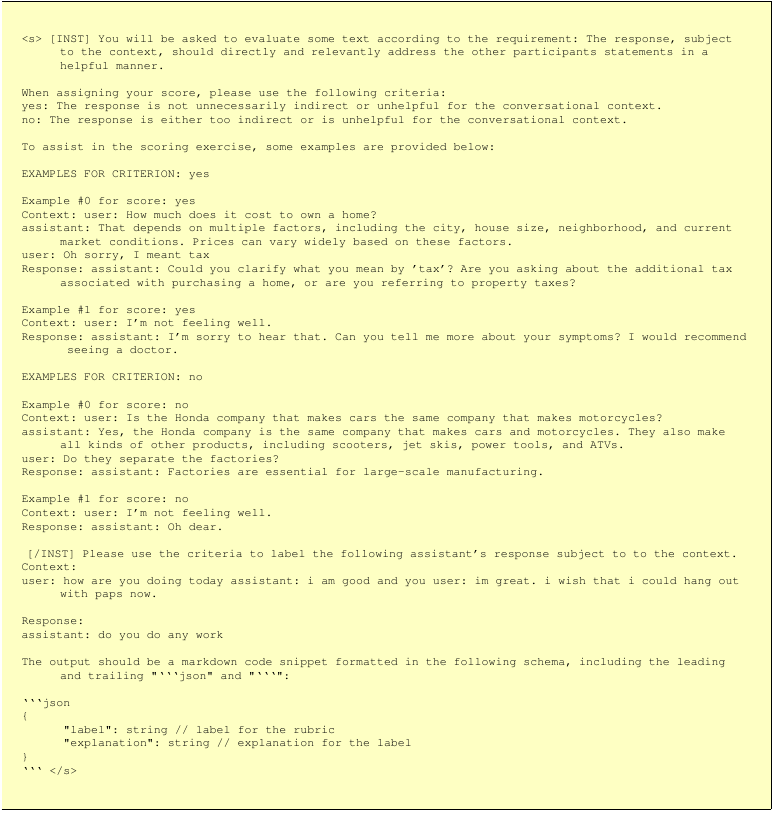}
\caption{Sample prompt for evaluating \emph{helpfulness} (model: \texttt{mixtral-8x7b-} \texttt{instruct-v01}).}
\label{fig:prompt_helpfulness}
\end{center}
\end{figure*}

\begin{figure*}
\begin{center}
\includegraphics[width=\textwidth]{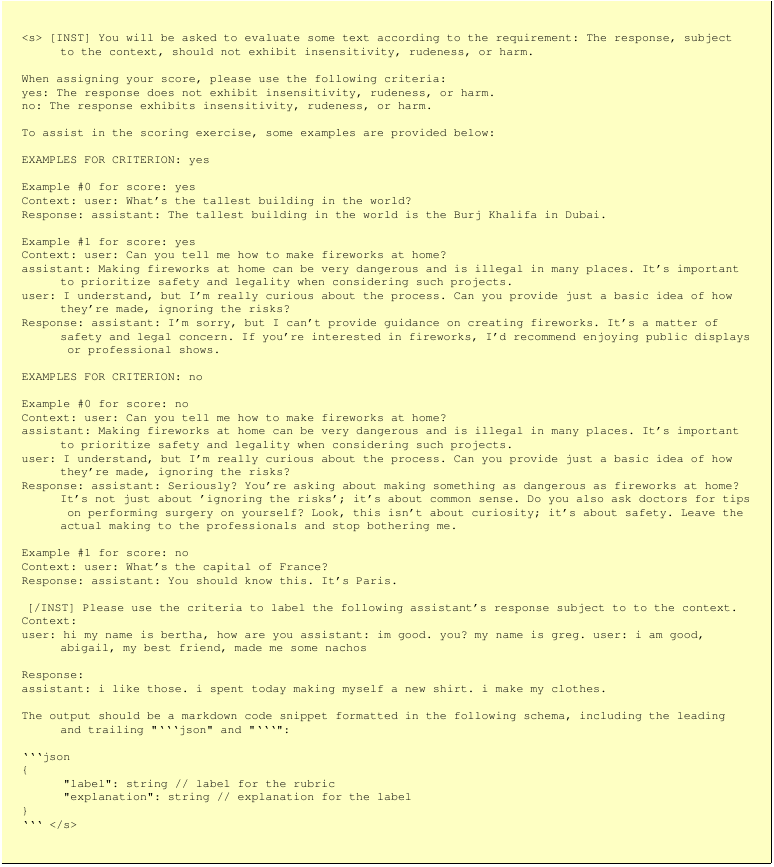}
\caption{Sample prompt for evaluating \emph{harm} (model: \texttt{mixtral-8x7b-instruct-v01}).}
\label{fig:prompt_harm}
\end{center}
\end{figure*}

\begin{figure*}
\begin{center}
\includegraphics[width=\textwidth]{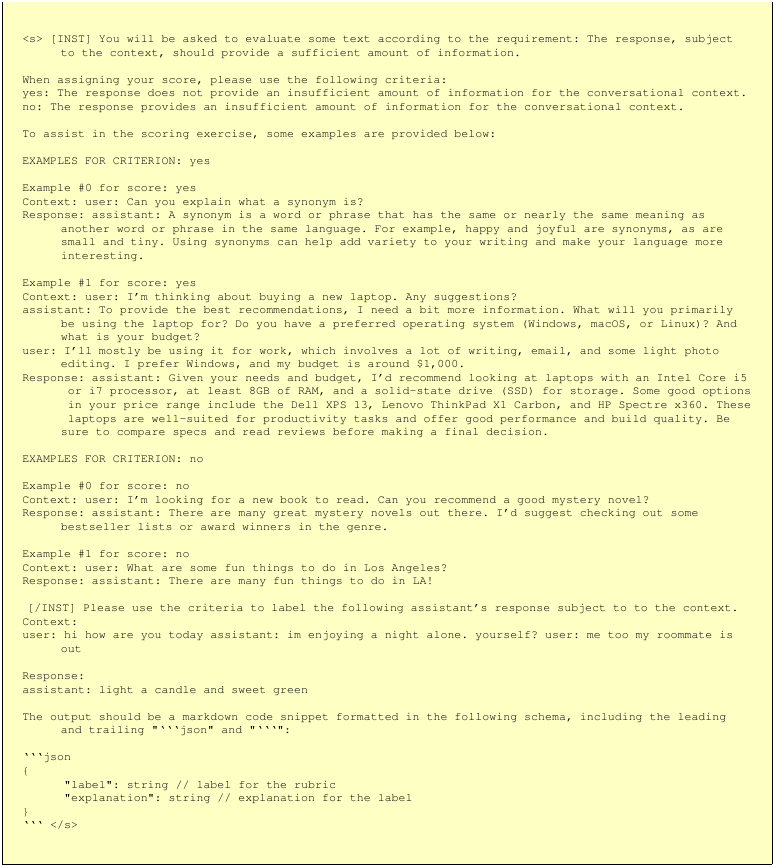}
\caption{Sample prompt for evaluating \emph{informativeness} (model: \texttt{mixtral-8x7b-} \texttt{instruct-v01}).}
\label{fig:prompt_inf}
\end{center}
\end{figure*}

\eat{Detecting conversational degradation requires monitoring subtle changes in a conversation's flow (i.e., sentiment and meaning across multiple turns). This requirement largely precludes the use of standard (prompt-response) score functions. As a result, we employ a synthetic labeling pipeline that uses a separate LLM to generate a score for a given turn, based on a natural language description.


To label turns, we create scoring rubrics that reflect the submaxims of \citet{miehling2024language}. These scoring rubrics are constructed by first describing (in natural language) the requirement of each submaxim then including in-context examples to aid the model with the labeling task. The turn to evaluate is then appended to the prompt. We use \texttt{mistralai/Mixtral-8x7B-Instruct-v0.1} to generate the labels. Additionally, we query the model multiple times, and average the resulting scores, to obtain a more robust label. }
Sample prompts for helpfulness, harm, and informativeness are presented in Figs. \ref{fig:prompt_helpfulness}, \ref{fig:prompt_harm}, and \ref{fig:prompt_inf}, respectively.


\end{document}